\documentclass{article}

\usepackage{microtype}
\usepackage{graphicx}
\usepackage{subcaption}
\usepackage{booktabs} 

\usepackage{hyperref}
\usepackage{balance}

\usepackage{tikz}
\usetikzlibrary{arrows.meta, positioning, decorations.pathreplacing}

\usepackage[preprint]{icml2026}

\usepackage{amsmath}
\usepackage{amssymb}
\usepackage{mathtools}
\usepackage{amsthm}

\usepackage[capitalize,noabbrev]{cleveref}

\theoremstyle{plain}
\newtheorem{theorem}{Theorem}[section]

\theoremstyle{definition}
\newtheorem{definition}[theorem]{Definition}

\theoremstyle{remark}

\usepackage[textsize=tiny]{todonotes}

\icmltitlerunning{What Do Learned Measurement Functions Measure?}

\begin{document}

\twocolumn[
\icmltitle{What Do Learned Models Measure?}

  \icmlsetsymbol{equal}{*}

  \begin{icmlauthorlist}
    \icmlauthor{Indr\.e \v{Z}liobait\.e}{yyy}
  \end{icmlauthorlist}

  \icmlaffiliation{yyy}{Department of Computer Science, University of Helsinki, Finland}

  \icmlcorrespondingauthor{Indr\.e \v{Z}liobait\.e}{indre.zliobaite@helsinki.fi}

  \icmlkeywords{Machine Learning,AI for science}

  \vskip 0.3in
]



\printAffiliationsAndNotice{}  

\begin{abstract}
In many scientific and data-driven applications, machine learning models are increasingly used as measurement instruments, rather than merely as predictors of predefined labels. When the measurement function is learned from data, the mapping from observations to quantities is determined implicitly by the training distribution and inductive biases, allowing multiple inequivalent mappings to satisfy standard predictive evaluation criteria. We formalize learned measurement functions as a distinct focus of evaluation and introduce measurement stability, a property capturing invariance of the measured quantity across admissible realizations of the learning process and across contexts. We show that standard evaluation criteria in machine learning, including generalization error, calibration, and robustness, do not guarantee measurement stability. Through a real-world case study, we show that models with comparable predictive performance can implement systematically inequivalent measurement functions, with distribution shift providing a concrete illustration of this failure. Taken together, our results highlight a limitation of existing evaluation frameworks in settings where learned model outputs are identified as measurements, motivating the need for an additional evaluative dimension.
\end{abstract}

\section{Introduction}
\label{sec:intro}

Machine learning models are increasingly used as components of measurement systems, rather than solely as predictors of predefined labels. In this role, learned models map raw observational data—such as sensor readings, images, or signals—to quantities that are treated as measurements and subsequently interpreted, compared, or acted upon.
When a learned model is used as a measurement instrument, it is implicitly expected to implement a mapping from observations to quantities that remains stable across conditions under which the measured quantity itself is preserved.

Traditionally, measurement instruments are defined by fixed, physically motivated transformations determined through explicit design and calibration. When measurement functions are instead learned from data, the mapping arises indirectly through modeling choices---such as the training data, model class, loss function, and optimization procedure---rather than through physical specification of the measured relationship.
As a result, learned models with comparable predictive performance can implement different mappings from observations to the same measured quantity.

Standard evaluation criteria in machine learning---such as generalization error, calibration, and robustness---assess predictive reliability.
They are well suited to predictive settings, but insufficient when learned models are used as measurement functions.
Predictive generalization under distribution shift does not guarantee that a learned model continues to measure the same quantity across contexts.

To motivate this distinction, consider thermometers whose sensing and calibration are implemented by learned models. Two such devices may each satisfy standard predictive validation yet exhibit systematic disagreement that varies with temperature. Such structured disagreement undermines the claim that either reliably measures temperature, indicating that the devices implement different measurement procedures despite low predictive error. In physical measurement, disagreement that depends on the quantity being measured is treated as measurement failure rather than noise, and agreement across admissible measurement procedures is required for the target quantity to be well-defined.

This observation highlights a gap in current evaluation methodology for learned systems.
When outputs of learned models are identified as measurements, the central evaluative question is not only whether predictions are accurate, but whether they measure what they are meant to measure.

In this paper, we argue that learned measurement functions constitute a distinct class of learning objects that are not adequately captured by existing evaluation frameworks.
We formalize the notion of a learned measurement function and introduce measurement stability, a property capturing invariance of the inferred quantity across admissible realizations of the learning process and across contexts in which that quantity is preserved.
We then motivate this problem through a real-world case study, where a learned model is used to measure temperature from sensory data. 

\textbf{This paper makes the following contributions}: 
(1) we formalize learned measurement functions as a distinct object of evaluation, separate from predictors assessed solely
by accuracy with respect to fixed labels;
(2) we introduce measurement stability as an evaluative property capturing invariance of the measured quantity across admissible model
realizations and deployment contexts;
(3) we provide real-world empirical evidence that predictive equivalence can mask systematic divergence in learned
measurements, particularly under distribution shift.

Our study lays the groundwork for evaluating learned models based on whether they measure the same quantity across contexts, making it possible to assess when they can be meaningfully interpreted as scientific instruments.

\section{Background and Related Work}
\label{sec:background}

Machine learning is traditionally formulated as the problem of learning a function that maps observations to predefined target variables specified independently of the learning process.
Evaluation criteria are correspondingly designed to assess how reliably a learned model predicts this fixed target under varying conditions.
Here we focus on an increasingly common setting \cite{Wang23}, in which learned models are identified as measurement instruments and mappings from observations to quantities that are subsequently treated as scientific measurements.

\textbf{Predictive evaluation and robustness.}
Standard machine learning evaluation frameworks emphasize generalization, calibration, and robustness as indicators of predictive reliability \cite{Shalev14,Guo17}.
Related work on robustness and distribution shift studies how predictive performance degrades under changes in the input distribution \cite{Ovadia19,Taori20}.
These approaches assume that the target quantity is fixed, and they assess stability relative to prediction error.
Our focus is complementary: even when predictive performance remains stable under standard evaluations, the learned mapping from observations to quantities of interest may change in systematic ways.
The proposed measurement instability therefore represents a failure mode that is not captured by standard predictive evaluation criteria.

\textbf{Algorithmic stability and identifiability.}
Several learning-theoretic frameworks seek predictors whose behavior is invariant across environments or perturbations, including algorithmic stability, invariant risk minimization, and distributionally robust optimization \cite{Bousquet02,Hardt16,Arjovsky19,Sagawa20}. Algorithmic stability ensures that small changes in training data lead to small changes in loss or predictions, but it does not guarantee uniqueness of the learned function or semantic equivalence across predictors. Invariant risk minimization seeks predictors whose optimality is invariant across environments; however, multiple statistically invariant predictors may still implement distinct measurement functions, particularly when environments are limited, proxies are correlated, or the task is underspecified. Distributionally robust optimization minimizes worst-case risk, but any minimizer achieving the same worst-case risk is admissible, and thus this does not guarantee uniqueness of the solution, semantic consistency, or agreement across minimizers. Taken together, these approaches constrain predictive behavior but do not, in general, ensure a unique mapping from observations to the measured quantity under a given training and evaluation procedure. We address this gap by asking when predictors that satisfy standard evaluation criteria can also be interpreted as measurements.

\textbf{Representation learning and identifiability.}
A large body of work in representation learning studies how to learn internal features that are predictive, invariant to nuisance variation, or identifiable up to a class of transformations \cite{Bengio13}. These approaches typically aim to recover representations that are sufficient for downstream tasks, and often explicitly tolerate non-uniqueness up to rotations, permutations, or other symmetries that preserve task performance. While such non-uniqueness is often benign for prediction, it can be consequential when learned representations are used as measurement instruments. Even when a target quantity is identifiable and predictive performance is optimal, measurements derived from learned representations may remain unstable across training realizations unless the learning objective explicitly constrains them. Our work complements representation learning by focusing not on learning a particular representation, but on evaluating the stability of measurements derived from learned models.

\textbf{Causality and invariance.}
A distinct notion of invariance arises in causal modeling, where stability is defined with respect to interventions on the data-generating process \cite{Pearl09,Peters17,Scholkopf21}. Causal modeling seeks relationships that persist under intervention, and causal invariance has been proposed as a principled basis for out-of-distribution generalization and scientific interpretation. Our notion of measurement stability is related but distinct. Causal invariance concerns stability of relationships under interventions on the data-generating process, whereas our measurement stability concerns invariance of the quantity being measured across admissible realizations of a learning process, holding fixed the interpretation of the target quantity. Even when a learned predictor aligns with a causal relationship, multiple admissible learning realizations may still lead to inequivalent measurement functions. In particular, causal assumptions may restrict which variables are relevant for prediction, but they do not by themselves guarantee uniqueness in the learning settings we consider. Measurement stability therefore addresses a complementary source of ambiguity in learned measurement settings.

\textbf{Underspecification.}
Recent work has highlighted that modern machine learning setups can be underspecified \cite{Damour20}: many models can achieve comparable predictive performance while differing in internal structure or downstream behavior.
Related issues arise in proxy objectives, or learned evaluators, which are assessed by agreement with held-out labels \cite{Amodei16,Gehrmann21}.
Our work builds on these insights by focusing on the consequences of underspecification.
Rather than asking whether a learned function predicts a target accurately, we ask whether different admissible learned mappings agree on what quantity is being measured.

\textbf{Scientific measurement validity.}
In scientific measurement, a central concern has long been whether an instrument measures a well-defined construct rather than merely producing repeatable outputs \cite{Stevens46,Hand04}.
Construct validity emphasizes agreement across admissible measurement procedures and treats systematic disagreement as evidence of measurement failure rather than noise.
We draw on this perspective to motivate learned measurement functions as objects of evaluation.
Our goal is to formalize an additional evaluative dimension—beyond standard machine learning criteria—that becomes necessary when learned models are interpreted as measurement instruments.

\section{Learned Measurement Functions}
\label{sec:measurement}

We formalize learned measurement functions and distinguish them from predictors
evaluated solely by predictive accuracy. This distinction is central to
understanding why standard evaluation criteria can fail when learned models are
used as components of measurement systems.

\subsection{Measurement Functions vs.\ Predictors}
\label{sec:measurement_vs_prediction}

In standard supervised learning, a model is trained to approximate a predictor
$f : X \rightarrow Y$,
where the target variable $Y$ is specified independently of the learning process.
Evaluation typically focuses on predictive accuracy, generalization, and robustness
with respect to a fixed loss function.

When learned models are used for measurement, the
learned function defines how raw observations are mapped to a numerical
representation that is subsequently treated as a measurement of an underlying
quantity of interest. We refer to such mappings as \emph{learned measurement
functions}.

\begin{definition}[Learned Measurement Function]
Let $\mathcal{Z}$ denote a space of possible quantities of interest, and
let $z \in \mathcal{Z}$ denote a particular quantity to be measured. Let
$\mathcal{C}_z$ denote a set of contexts over which $z$ is assumed to remain
invariant in its interpretation. A learned measurement function is a mapping
$m_\theta : X \times \mathcal{C}_z \rightarrow \mathbb{R}$,
parameterized by learned parameters $\theta$, where the output $m_\theta(x,c)$ is
interpreted as a measurement of the latent quantity $z$.
\end{definition}

In this setting, the meaning of the measured quantity is not fixed a priori.
There is no externally specified rule that uniquely determines what the model output represents as a measurement; instead, this interpretation is shaped implicitly by the training data, model class, inductive biases, and optimization procedure.
Consequently, the meaning of the output (e.g., “surface temperature”, “blood pressure”, "energy momentum")  emerges from the learning process.
This contrasts with physical measurement instruments, such as thermometers, for which the measured quantity is defined independently of any particular realization of the instrument.
As a result, two learned models with comparable predictive performance may implement different learned measurement functions, even when evaluated under identical predictive criteria.
The remaining question is why learned measurement settings admit multiple admissible mappings, whereas classical measurement does not. Understanding how such non-uniqueness arises requires examining the role of inductive bias in learned measurement functions.

\subsection{Non-Uniqueness and Inductive Bias}
\label{sec:nonuniqueness}

Classical physical measurement provides a useful point of contrast.
Instruments for measuring temperature, for example, implement physically grounded mappings from system states to numerical values. Different instruments may rely on distinct mechanisms—such as thermal expansion, electrical resistance, or infrared radiation—but they are treated as measuring the same quantity insofar as they agree, up to calibration and noise. The quantity being measured is defined not by the internal construction of any particular instrument, but by its invariance across admissible realizations.

By an admissible realization, we mean a variation of the measurement procedure---such as a different instrument, model instantiation, training run, or data sample---that is intended to measure the same underlying quantity and is therefore treated as a legitimate alternative rather than a change in what is being measured. By contrast, changes that redefine the target quantity or alter the measurement goal---for example, measuring surface temperature instead of ambient temperature---are not admissible realizations.

Learned measurement functions differ from prescribed measurement instruments in that the mapping from observations to quantities is not specified explicitly, but inferred from data. Learning objectives typically impose only risk-based constraints, admitting a set of risk-equivalent solutions. Inductive and training data choices determine which mapping is selected as the measurement function.

As a result, the same training objective and evaluation protocol may admit multiple learned measurement functions that implement different relationships between observations and the quantity of interest. 
While such non-uniqueness is often inconsequential for prediction tasks, it becomes central when models are used as measurement instruments, as admissible learned measurement functions may then disagree systematically across contexts.
Systematic disagreement is problematic because it reflects ambiguity in what is being measured rather than random noise or uncertainty.

\subsection{Measurement as an Evaluation Object}
\label{sec:measurement_object}

These observations motivate treating learned measurement functions as a distinct object of evaluation.
Rather than evaluating a single trained model in isolation, we propose to consider the set of learned measurement functions that are admissible under a given training and evaluation workflow.

From this perspective, the relevant evaluation question is not only whether a learned model predicts accurately, but whether different admissible learned measurement functions agree on what quantity is being measured.
If models that are equivalent under standard predictive criteria implement inequivalent measurement functions, then the measurement itself is underspecified.
This reframing highlights a limitation of standard evaluation practice.

\section{Measurement Stability}
\label{sec:stability}

We introduce measurement stability, a property that characterizes when learned measurement functions support reliable measurement of a well-defined quantity.

\subsection{Definition of Measurement Stability}
\label{sec:stability_definition}

Informally, a learned measurement is stable if different admissible realizations of the learning process produce measurements that agree when applied to the same underlying observations across contexts in which the underlying quantity of interest is assumed to remain unchanged.
Measurement stability therefore concerns invariance of what is being measured, rather than consistency of predictions produced by a particular model instance.

Formally, measurement stability is defined as follows.
\begin{definition}[Measurement Stability]
\label{def:measurement_stability}
Let $\mathcal{Z}$ denote a space of possible quantities of interest, and let
$z \in \mathcal{Z}$ denote a particular quantity to be measured.
Let $\mathcal{C}_z$ be a set of contexts over which $z$ is assumed to remain
invariant, in the sense that variation in $c \in \mathcal{C}_z$ does not alter
which quantity is being measured.
Let $\mathcal{O}$ denote a space of underlying observations.
Let $\mathcal{L}$ be a learning process whose admissible
realizations induce measurement functions
\[
m : X_m \times \mathcal{C}_z \to \mathbb{R},
\]
where $X_m$ denotes a representation or feature space that may depend on the
learning realization.

Let $\phi_m : \mathcal{O} \to X_m$ denote the observation-to-representation mapping
associated with $m$, and let $\mathcal{M}_z$ denote the set of measurement functions
induced by admissible realizations of $\mathcal{L}$ under these assumptions.
The measurement of $z$ is \emph{stable} if
\[
\begin{aligned}
\forall m_1, m_2 \in \mathcal{M}_z,\;
\forall o \in \mathcal{O},\;
\forall c \in \mathcal{C}_z:\quad \\
m_1(\phi_{m_1}(o), c) \approx m_2(\phi_{m_2}(o), c)
\end{aligned}
\]
where $\approx$ denotes equivalence under the admissible transformations of the
measurement scale associated with $z$.
\end{definition}

Measurement stability concerns agreement of measurements, not agreement of representations; different learning realizations may use different feature spaces.

Measurement instability arises when multiple admissible learned measurement functions satisfy the same predictive objective but implement different inferential mappings from observations to the quantity of interest.

This definition makes explicit that measurement stability is a relational property across sets of admissible models and contexts, rather than a property of any single trained model.
Evaluating stability requires comparing outputs across admissible realizations of the learning process or across contexts in which the underlying quantity is preserved.

\subsection{Contexts and Admissible Realizations}
\label{sec:contexts}

The notion of an admissible realizations includes transformations under which the underlying quantity of interest is preserved.
Examples include changes in data collection conditions, sensor characteristics, preprocessing, or deployment environments that do not alter the quantity being measured.

In physical measurement, admissible transformations are fixed by domain knowledge and calibration standards.
In learned measurement settings, admissibility is implicit and determined by modeling assumptions and data availability.
Measurement stability therefore requires making explicit which variations are considered measurement-preserving and assessing agreement across them.

We define an admissible realization as any instantiation of the learning procedure that is consistent with the stated training protocol, including the same data-generating distribution and learning objective, but differing in elements not fixed by predictive validation.
We intentionally define admissible realizations at this level of abstraction to capture the underdetermination induced by predictive objectives themselves, rather than artifacts of a particular optimizer or architecture.

Admissible realizations need not correspond to worst-case perturbations.
Rather, they represent variations that are treated as routine in practice, such as retraining with new data, deploying across institutions, or updating model implementations.
Stability concerns whether learned measurement functions agree across such variations, not whether predictions are invariant under arbitrary input perturbations.


\subsection{Why Standard Evaluation Criteria Are Insufficient}
\label{sec:stability_vs_standard}

Standard evaluation criteria in machine learning assess predictive reliability, but they do not guarantee measurement stability. When model outputs are interpreted as measurements, satisfying standard evaluation criteria remains necessary but is not sufficient.

\textbf{Generalization}
bounds expected prediction error with respect to a fixed target variable.
Generalization does not restrict which predictor is selected among multiple risk-equivalent solutions.
As a result, two models may generalize equally well while implementing different mappings from observations to the quantity of interest, leading to unstable measurement despite strong generalization performance.

\textbf{Calibration}
assesses whether predicted values or confidence estimates align with empirical error frequencies.
Well-calibrated models can nevertheless disagree systematically on the value of a measured quantity, particularly under distribution shift or retraining.
Calibration therefore characterizes internal consistency of predictions, not agreement across admissible measurement functions.

\textbf{Robustness} is the closest to what we are aiming at, and yet distinct.
Robustness evaluations examine sensitivity of predictive performance to distributional changes.
Robustness fixes a learned model and asks whether its outputs change under perturbations of inputs, data, or environments. Measurement stability inverts this quantification: it fixes the underlying quantity of interest and asks whether different admissible learning realizations induce the same measurement of that quantity across contexts.
A learned measurement function may be robust in this sense while still changing what quantity is effectively being measured across contexts.
Robustness thus does not, by itself, rule out systematic divergence among admissible learned measurement functions.

More formally, robustness studies invariance of predictions for a fixed learned model,\\
$\forall m \;\forall \delta \in \Delta:\quad
m(x) \approx m(x+\delta)$, where $\delta$ ranges over admissible perturbations of inputs, data, or environments.

Measurement stability instead fixes a chosen quantity of interest $z \in \mathcal{Z}$
and studies invariance of the induced measurement across admissible realizations of
the learning process. Let
$\mathcal{O}$ denote a space of underlying observations and let
$m(\phi_m(o), c)$ denote the measurement assigned by realization $m$ to observation
$o$ in context $c$. Measurement stability requires
$\forall z \in \mathcal{Z},\;
\forall m_1, m_2 \in \mathcal{M}_z,\;
\forall o \in \mathcal{O},\;
\forall c \in \mathcal{C}_z:\quad \\
m_1(\phi_{m_1}(o), c) \approx m_2(\phi_{m_2}(o), c)$
holding the underlying quantity of interest $z$ fixed.

Taken together, generalization, calibration, and robustness evaluate the reliability of predictions produced by a single model realization with respect to a fixed target variable. They ask whether a model predicts accurately, consistently, and reliably under various conditions, assuming that the predicted quantity is already well defined.

Measurement stability, by contrast, concerns whether different admissible realizations of the learning process—such as retraining, model variation, or deployment across contexts—agree on how observations map to the quantity of interest. It evaluates whether the measured quantity itself remains invariant across admissible contexts, rather than whether any particular model predicts well.

As a result, a learned model can generalize well, be well calibrated, and be robust to perturbations, yet still implement a measurement function whose outputs differ systematically from those of other admissible models. Conversely, agreement across admissible models does not guarantee high predictive accuracy. Because these properties vary independently, measurement stability is orthogonal to standard predictive evaluation criteria rather than implied by them.

\section{Empirical Evidence of Failure}
\label{sec:empirical}

This section provides empirical evidence that measurement instability is not merely a theoretical concern. In a real-world setting, we show that multiple models achieving comparable generalization nevertheless implement systematically different measurement functions. 
These differences persist under realistic distribution shifts and lead to divergent downstream conclusions, directly validating the concerns raised in the previous sections. 

Our goal is not to introduce a new modeling
approach, but to audit a standard learning and evaluation procedure 
and show that it can certify multiple admissible learned measurement functions
for a fixed quantity of interest that fail the measurement stability
condition, despite satisfying standard evaluation criteria. 
In such a case, the predictions cannot be interpreted as measurements of the
quantity of interest. 

\subsection{Experimental Setup}
\label{sec:setup}

Consider a sensor-based measurement task in which learned models are used to map
observational inputs to numerical values that are subsequently treated as
measurements of an underlying quantity of interest.
We use the UCI Air Quality dataset \cite{UCI,Vito08}, which contains time-stamped readings from
multiple chemical sensors ($X$) along with environmental variables and temperature serves as the quantity of interest
($z \in \mathcal{Z}$).

We train two linear models to infer temperature using different, partially overlapping subsets of sensors.
For the demonstration purpose the sensor subsets are chosen such that 
both models support accurate prediction of temperature, but rely on different observational proxies.
This reflects a common real-world scenario in which different instruments, sensor
configurations, or preprocessing pipelines are used to measure the same quantity.

We train both models on an initial time window and evaluate on a subsequent time
window after a gap, inducing a natural temporal distribution shift.
This shift is expected to alter the relationship between sensor readings and
temperature while preserving the interpretation of the quantity being measured.
Accordingly, the two trained models can be treated as admissible realizations of a
learning process intended to measure the same quantity $z$, rather than as solutions
to distinct prediction problems.

Code for replication is provided in the Appendix.

\subsection{Performance under Standard Evaluation Criteria}
\label{sec:criteria}

We first assess how two learned models perform under standard evaluation criteria of machine learned models --- generalization, calibration and robustness.

\begin{figure*}[t]
    \centering
    \includegraphics[width=0.9\textwidth]{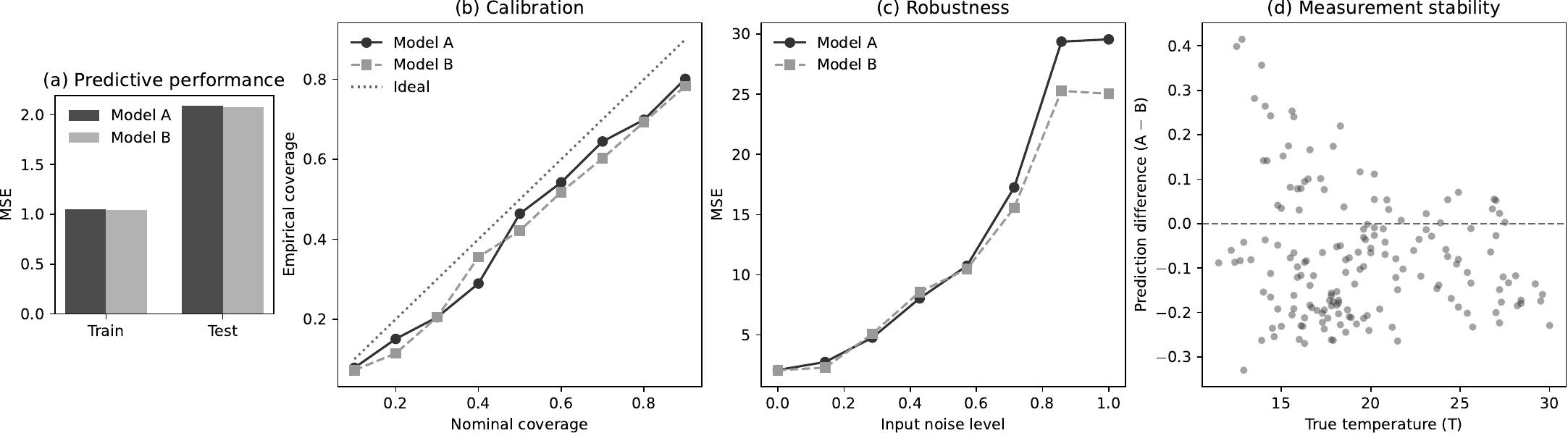}
    \caption{
    \textbf{Evaluation of predictive accuracy and measurement stability.}
    (a) Predictive performance of Model~A and Model~B on training and held-out test data. MSE --- mean squared error.
    (b) Calibration curves under temporal distribution shift comparing nominal and empirical coverage.
    (c) Robustness under increasing input noise.
    (d) Measurement stability: prediction disagreement between two models trained on different sensor subsets as a function of true temperature.
    }
    \label{fig:realdata}
\end{figure*}

Figure~\ref{fig:realdata}(a) reports predictive performance on training data and on
temporally shifted test data.
The two models achieve comparable mean squared error both in-distribution and under
temporal shift.
Under standard predictive evaluation, the models would therefore be considered
interchangeable.

Next we evaluate predictive calibration for both models under temporal distribution
shift.
Following standard practice for regression, we assume a Gaussian predictive model
with variance estimated from training residuals.
For a range of nominal confidence levels, symmetric prediction intervals are
constructed and empirical coverage is computed on held-out test data, allowing
calibration to be assessed by comparing nominal and empirical coverage.

Figure~\ref{fig:realdata}(b) reports the resulting calibration curves.
Both models exhibit broadly comparable calibration: empirical coverage
tracks nominal coverage reasonably well across confidence levels, with no
systematic divergence between models.
While calibration is not perfect, neither model exhibits a clear calibration
failure under this evaluation.

Calibration evaluates agreement between predictive uncertainty and
empirical error relative to a fixed target variable.
It does not assess whether different admissible learning realizations induce
equivalent measurements of the underlying quantity.
Calibration therefore provides no guarantee of measurement stability.

We additionally evaluate predictive robustness by measuring performance degradation
under increasing input perturbations.
Specifically, we add zero-mean Gaussian noise of increasing variance to standardized
test inputs and measure mean squared error as a function of noise magnitude.

Figure~\ref{fig:realdata}(c) shows predictive error as a function of input noise level
for both models.
The two models exhibit similar degradation trends: predictive error increases
smoothly with noise magnitude, and neither model displays a qualitatively distinct
robustness failure.

As with calibration, robustness evaluation concerns invariance of predictive
performance for a fixed model under perturbations.
It does not constrain agreement across admissible realizations of the learning
process, and therefore does not guarantee measurement stability.

These analyses are included to establish that the later observed measurement divergence cannot be
attributed to failures of predictive reliability under standard criteria.
Taken together, these results indicate that both models would be deemed acceptable
under standard evaluation protocols.

\subsection{Assessing Measurement Stability}
\label{sec:divergence}

We next assess the measurement stability of the learned measurements of the
quantity of interest. Note that measurement stability is not a property of individual model instances,
but of the measurements they induce across admissible realizations of the learning
process for a fixed quantity of interest.
Here, we are thus assessing whether the outputs of either or both of these models
can be interpreted as valid measurements of the quantity of interest.

Figure~\ref{fig:realdata}(d) visualizes disagreement between the learned measurement
functions when applied to the same test observations.
While the models achieve comparable predictive accuracy, their outputs diverge
in a structured, state-dependent manner as a function of the inferred temperature.

This disagreement is not consistent with zero-mean noise.
Instead, the magnitude and direction of disagreement vary systematically along with the target, indicating that the models implement different mappings from
observations to numerical measurements of temperature.
Under Definition~\ref{def:measurement_stability}, this is a failure, as the measured
quantity systematically depends on the realization of the learning process.

When two admissible realizations disagree, the framework does not license us to declare one valid and the other invalid.
Validity as a measurement is not assessed model-by-model.
A failure of stability implies that the measurement is not well-defined, not that one model is underperforming.

\subsection{Stable Predictions, Unstable Measurements}
\label{sec:shift}

Temporal distribution shift provides a particularly clear diagnostic for measurement
instability.
Here, sensor behavior evolves over time while the interpretation of the
quantity of interest—temperature—remains fixed.
Temporal shift isolates differences in how admissible learned
measurement functions respond to this change in operating context.

Under temporal shift, predictive performance remains stable for both models, and
robustness analysis checks indicate similar degradation behavior under input perturbations.
However, divergence between the learned measurement functions becomes 
pronounced, as shown in Figure~\ref{fig:realdata}(d).

Crucially, the shift does not merely degrade predictive accuracy.
Instead, it reveals that different admissible realizations of the learning process
assign systematically different measurements to the same preserved quantity.
This state-dependent divergence is characteristic of measurement instability and is
not detected by standard evaluations of generalization, calibration, or robustness.

Although the learned models rely on different feature subsets, this does not by itself imply non-identifiability of the target quantity. The estimand remains well defined and identifiable from the data; what is underdetermined is the measurement mechanism implemented by the learning procedure. As a result, predictors with similar in-distribution and standard out-of-distribution performance can nonetheless exhibit structural disagreement under distribution shift.

Training models on different feature subsets yields predictors that are empirically indistinguishable under standard evaluation yet diverge systematically under distribution shift. This divergence is problematic as it implies that the measured value depends on the choice of measurement mechanism rather than on the system being measured. In scientific settings, such dependence undermines instrument interchangeability and renders the measurement ill-defined, even when predictive performance appears adequate.

Taken together, these results demonstrate a violation of the measurement stability
condition of Definition~\ref{def:measurement_stability}.
Despite satisfying conventional criteria for generalization, calibration, and
robustness, the learned measurement functions disagree systematically when applied
to the same underlying observations across invariant contexts.

Standard evaluation assesses the predictive reliability of individual trained
models. It implicitly treats models as interchangeable whenever they satisfy the same predictive criteria.
While appropriate for prediction, this assumption is problematic when models are
used as measurement instruments, where agreement across admissible realizations is
a prerequisite for stable measurement.

Even if satisfied, measurement stability is not a substitute for predictive accuracy,
robustness, or calibration.
Valid measurements must still be accurate, robust to perturbations, and
appropriately calibrated.
Measurement stability instead constitutes an additional and independent evaluative
dimension, required to ensure that learned model outputs admit a coherent
interpretation as measurements of a fixed quantity across admissible realizations
of the learning process.

\section{Implications for Evaluation Practice}
\label{sec:implications}

Our results have direct implications for how learned systems should be evaluated
when their outputs are interpreted as measurements rather than as predictions.

\textbf{What standard evaluation cannot certify.}
Standard evaluation can certify that a particular trained model predicts reliably
with respect to a fixed target variable.
It cannot, in general, certify that the learned mapping constitutes a stable
measurement of a well-defined quantity.
No evaluation that inspects a single trained model in isolation can establish
measurement stability, which is inherently relational, concerning agreement across
admissible realizations of the learning process and across contexts.

\textbf{When learned measurements are risky.}
Measurement instability is most likely in settings where the mapping from observations
to the quantity of interest is weakly constrained, such as when quantities are latent
or abstract, proxies are indirect, or multiple representations achieve comparable
predictive performance.
High-dimensional sensing, representation learning, and scientific inference are
therefore particularly susceptible, as are operational settings involving retraining,
model replacement, or deployment across institutions.
Notably, this risk can persist even when predictive performance shows no degradation.

\textbf{Implications for evaluation practice.}
When learned systems are used as measurement instruments, several concrete
implications for evaluation practice follow.
First, variability across training realizations should be treated as signal rather
than noise.
Practitioners should intentionally train multiple admissible realizations of the
learning process and evaluate agreement between the resulting measurement functions.

Second, measurement stability should be evaluated alongside predictive performance
through task-appropriate measures of between-model disagreement.
Structural disagreement indicates that the learned quantity is not well-defined for
measurement, even when predictive metrics are satisfied.

Third, model selection should explicitly consider stability–performance tradeoffs.
When multiple models achieve comparable predictive performance, preference should be
given to models whose induced measurements are more stable across admissible
realizations and contexts, even at the cost of modest reductions in accuracy.

\textbf{Implications for interpretability.}
Our results also clarify how robustness can be interpreted when used to
justify interpretability claims.
Robustness to distribution shift does not imply semantic stability: a model may be robust in its predictions while the quantities it appears
to measure vary across realizations or environments.
Consequently, robustness alone is insufficient to support claims that learned
features correspond to stable, well-defined target quantities.

This limitation extends to post-hoc interpretability methods, whose conclusions are
meaningful only to the extent that the underlying measurements are stable.
Measurement stability therefore functions as a diagnostic precondition for
interpretability, analogous to identifiability in causal inference.


\section{Limitations and Outlook}
\label{sec:limitations}

This paper adopts a diagnostic perspective on learned measurement, focusing on how
learned systems function as measurement instruments.
We characterize a failure mode that is not captured by current evaluation criteria
and show how it can arise under standard learning and deployment conditions.
Our goal is to make this gap in evaluation practice explicit and to clarify
the conditions under which it becomes relevant, thereby providing groundwork for
future methodological work on learned measurement.

Measurement stability is not directly observable from data produced by a single trained model. Stability concerns invariance of the inferred quantity across admissible realizations or deployment contexts, and is therefore a property of the learning setup; it diagnoses whether the learned measurement is well-defined. Certifying stability consequently requires access to multiple admissible realizations of the learning process, or to controlled variations that probe agreement across them. A single model thus cannot be optimized for measurement stability in isolation, whereas a learning setup can be designed to promote it. Operationalizing statistical guarantees for such setups is left as an open problem outside the scope of the present paper.

Measurement instability can possibly arise from two distinct sources: (i) causal non-identifiability of the target quantity, in which case no learning procedure can recover a unique measurement even with infinite data, and (ii) epistemic underdetermination induced by the learning process, where multiple empirically equivalent measurement functions exist despite identifiability in principle. Our current framework is agnostic to this distinction: measurement stability is a property of the learned model relative to validation criteria, not a diagnosis of its cause. Our central message is that standard predictive evaluation obscures this distinction.

Finally, this work focuses on settings in which learned model outputs are interpreted as measurements in downstream analysis. Models that fail to satisfy measurement stability may still be highly effective for
prediction, ranking, or decision support, and our analysis does not call their use
in these roles into question.
Rather, we show that when measurement stability is absent, model outputs should be
understood as predictive proxies rather than as principled measurements, particularly
in scientific or other measurement-driven contexts.

Looking forward, our analysis points to several directions for future work.
One is to formalize modeling practices that
explicitly assess agreement across admissible realizations of the learning process.
Another is to delineate more precisely the boundaries of what learned
model evaluation can and cannot certify about measurement validity.
A further direction is to incorporate assumptions about admissible contexts and preserved quantities
directly into learning. 


\section{Conclusion}

We argued that when learned models are used as measurement instruments, predictive
reliability alone is insufficient for principled interpretation.
Measurement stability---agreement across admissible realizations of the learning
process---is a distinct requirement that is not captured by standard evaluation
criteria.
Recognizing this distinction clarifies the scope of current evaluation practice and
motivates the development of methods that explicitly support stable learned
measurement.




 \newpage

\section*{Impact Statement}

\balance 

Using machine learning models as measurement instruments introduces evaluative considerations that go beyond predictive performance. By formalizing the notion of measurement stability, this work highlights a structural limitation of standard evaluation practices when learned models are deployed in this role. The analysis is primarily conceptual and does not introduce new algorithms or applications. While the results do not pose direct societal risks, they aim to inform more careful evaluation and interpretation of learned models in scientific contexts, where the validity of measurement and the assignment of epistemic responsibility are critical. More broadly, the work clarifies conditions for the reliable integration of machine learning into scientific and analytical practices.

\bibliography{bib_ai_icml}
\bibliographystyle{icml2026}

\appendix
\onecolumn

\section{Reference Implementation}
\label{app:code}

The full code used to generate all experimental results and figures in this paper is
provided below.

The code is self-contained and operates on the UCI Air Quality dataset, which is
publicly available \cite{UCI,Vito08}.
No additional experimental details are required beyond those specified in the main
text.

{\tiny
\begin{verbatim}
import pandas as pd
import numpy as np
import matplotlib.pyplot as plt
import matplotlib.gridspec as gridspec

from sklearn.linear_model import LinearRegression
from sklearn.preprocessing import StandardScaler
from sklearn.metrics import mean_squared_error
from scipy.stats import norm

# ===Load and clean data===
df = pd.read_csv("AirQualityUCI.csv", sep=";", decimal=",")
df = df.loc[:, ~df.columns.str.contains("^Unnamed")]
df = df.replace(-200, np.nan)
df = df.dropna().reset_index(drop=True)

# ===Define learning task===
y = df["T"].values
sensors_shared = ["AH", "RH"]
sensors_A_only = ["PT08.S1(CO)", "PT08.S3(NOx)"]
sensors_B_only = ["PT08.S2(NMHC)", "PT08.S4(NO2)"]
sensors_A = sensors_shared + sensors_A_only
sensors_B = sensors_shared + sensors_B_only

# ===Temporal split, train and test sets===
n = len(df)
train_frac = 0.6
gap_frac = 0.2
train_end = int(train_frac * n)
gap_end = int((train_frac + gap_frac) * n)
df_train = df.iloc[:train_end]
df_test  = df.iloc[gap_end:]
y_train = df_train["T"].values
y_test  = df_test["T"].values
X_A_train = df_train[sensors_A].values
X_A_test  = df_test[sensors_A].values
X_B_train = df_train[sensors_B].values
X_B_test  = df_test[sensors_B].values

# ===Standardization===
scaler_A = StandardScaler()
scaler_B = StandardScaler()
X_A_train = scaler_A.fit_transform(X_A_train)
X_A_test  = scaler_A.transform(X_A_test)
X_B_train = scaler_B.fit_transform(X_B_train)
X_B_test  = scaler_B.transform(X_B_test)

# ===Train models===
f_A = LinearRegression().fit(X_A_train, y_train)
f_B = LinearRegression().fit(X_B_train, y_train)

# ===Predict===
pred_A_train = f_A.predict(X_A_train)
pred_B_train = f_B.predict(X_B_train)
pred_A_test = f_A.predict(X_A_test)
pred_B_test = f_B.predict(X_B_test)

# ===Measure performance===
mse_A_train = mean_squared_error(y_train, pred_A_train)
mse_B_train = mean_squared_error(y_train, pred_B_train)
mse_A_test = mean_squared_error(y_test, pred_A_test)
mse_B_test = mean_squared_error(y_test, pred_B_test)
print("Predictive MSE (train): A =", mse_A_train, "B =", mse_B_train)
print("Predictive MSE (test):  A =", mse_A_test,  "B =", mse_B_test)

# ===Measurement disagreement test===
disagreement_test = pred_A_test - pred_B_test

# ===Calibration analysis===
def calibration_curve_regression(y_true, y_pred, sigma, levels):
    coverages = []
    for alpha in levels:
        z = norm.ppf(1 - (1 - alpha) / 2)
        lower = y_pred - z * sigma
        upper = y_pred + z * sigma
        coverages.append(np.mean((y_true >= lower) & (y_true <= upper)))
    return np.array(coverages)

resid_A = y_train - pred_A_train
resid_B = y_train - pred_B_train
sigma_A = np.std(resid_A)
sigma_B = np.std(resid_B)
levels = np.linspace(0.1, 0.9, 9)
cov_A = calibration_curve_regression(y_test, pred_A_test, sigma_A, levels)
cov_B = calibration_curve_regression(y_test, pred_B_test, sigma_B, levels)

# ===Robustness analysis===
noise_levels = np.linspace(0.0, 1.0, 8)
mse_A_curve = []
mse_B_curve = []
for s in noise_levels:
    X_A_noisy = X_A_test + np.random.normal(0, s, X_A_test.shape)
    X_B_noisy = X_B_test + np.random.normal(0, s, X_B_test.shape)
    mse_A_curve.append(mean_squared_error(y_test, f_A.predict(X_A_noisy)))
    mse_B_curve.append(mean_squared_error(y_test, f_B.predict(X_B_noisy)))
    
# ===Average agreement test===
mean_disagreement = np.mean(disagreement_test)
std_disagreement = np.std(disagreement_test)
print("Average disagreement (A - B) on test set:", mean_disagreement)
print("Std. deviation of disagreement on test set:", std_disagreement)
relative_mean_disagreement = mean_disagreement / np.std(y_test)
print("Relative mean disagreement:", relative_mean_disagreement)

# ===Figure===
fig = plt.figure(
    figsize=(14.5, 4.6),
    constrained_layout=True
)
gs = gridspec.GridSpec(
    1, 4,
    figure=fig
)
ax_a = fig.add_subplot(gs[0, 0])
ax_b = fig.add_subplot(gs[0, 1])
ax_c = fig.add_subplot(gs[0, 2])
ax_d = fig.add_subplot(gs[0, 3])

labels = ["Train", "Test"]
x = np.arange(len(labels))
width = 0.35
ax_a.bar(x - width/2, [mse_A_train, mse_A_test], width,
         color="0.3", label="Model A")
ax_a.bar(x + width/2, [mse_B_train, mse_B_test], width,
         color="0.7", label="Model B")
ax_a.set_xticks(x)
ax_a.set_xticklabels(labels)
ax_a.set_ylabel("MSE")
ax_a.set_title("(a) Predictive performance")
ax_a.legend(frameon=False)
ax_a.set_box_aspect(1)

ax_b.plot(levels, cov_A, "o-", color="0.2", label="Model A")
ax_b.plot(levels, cov_B, "s--", color="0.6", label="Model B")
ax_b.plot(levels, levels, ":", color="0.4", label="Ideal")
ax_b.set_xlabel("Nominal coverage")
ax_b.set_ylabel("Empirical coverage")
ax_b.set_title("(b) Calibration")
ax_b.legend(frameon=False)
ax_b.set_box_aspect(1)

ax_c.plot(noise_levels, mse_A_curve, "o-", color="0.2", label="Model A")
ax_c.plot(noise_levels, mse_B_curve, "s--", color="0.6", label="Model B")
ax_c.set_xlabel("Input noise level")
ax_c.set_ylabel("MSE")
ax_c.set_title("(c) Robustness")
ax_c.legend(frameon=False)
ax_c.set_box_aspect(1)

ax_d.scatter(
    y_test,
    disagreement_test,
    s=24,
    color="0.2",
    alpha=0.45,
    edgecolors="none"
)
ax_d.axhline(0, linestyle="--", linewidth=1.2, color="0.4")
ax_d.set_xlabel("True temperature (T)")
ax_d.set_ylabel("Prediction difference (A - B)")
ax_d.set_title("(d) Measurement stability")
ax_d.set_box_aspect(1)

plt.savefig("fig_evaluation.pdf")
plt.close()
\end{verbatim}
}

\end{document}